\documentclass[a4paper]{jpconf}
\usepackage{graphicx}
\usepackage{amsmath}

\usepackage{bm}
\usepackage{amssymb}
\usepackage{multirow}
\usepackage{subfig}

\newcommand{\TC}{$T_{\rm C}$}
\usepackage[ruled, linesnumbered, boxed, vlined ]{algorithm2e} 
\usepackage{nth}

\usepackage{float}

\SetKw{KwBy}{by}
\SetKwInOut{Parameter}{Parameters}

\usepackage{color}

\begin{document}
\title{Ensemble learning reveals dissimilarity between rare-earth transition metal binary alloys with respect to the Curie temperature}
\author{Duong-Nguyen Nguyen$^1$, Tien-Lam Pham$^{1,2}$, Viet-Cuong Nguyen$^3$, Hiori Kino$^{2,4}$, Takashi Miyake$^{2,4,5}$, Hieu-Chi Dam$^{1,4,7}$ }

\address{$^{1}$Japan Advanced Institute of Science and Technology, 1-1 Asahidai, Nomi, Ishikawa 923-1292, Japan}
\address{$^{2}$ESICMM, NIMS, Tsukuba 305-0047, Japan}
\address{$^{3}$HPC Systems Inc., 3-9-15 Kaigan, Minato-ku, Tokyo 108-0022, Japan}
\address{$^{4}$Center for Materials Research by Information Integration, National Institute for Materials Science, 1-2-1 Sengen, Tsukuba, Ibaraki 305-0047, Japan}
\address{$^{6}$CD-FMat, AIST, 1-1-1 Umezono, Tsukuba 305-8568, Japan}
\address{$^{7}$JST, PRESTO, 4-1-8 Honcho, Kawaguchi, Saitama 332-0012, Japan}

\ead{dam@jaist.ac.jp}

\begin{abstract}
We propose a data-driven method to extract dissimilarity between materials, with respect to a given target physical property. The technique is based on an ensemble method with Kernel ridge regression as the predicting model; multiple random subset sampling of the materials is done to generate prediction models and the corresponding contributions of the reference training materials in detail. The distribution of the predicted values for each material can be approximated by a Gaussian mixture models. The reference training materials contributed to the prediction model that accurately predicts the physical property value of a specific material, are considered to be similar to that material, or vice versa.  Evaluations using synthesized data demonstrate that the proposed method can effectively measure the dissimilarity between data instances. An application of the analysis method on the data of Curie temperature ({\TC}) of binary 3$d$ transition metal- 4$f$ rare-earth binary alloys also reveals meaningful results on the relations between the materials. The proposed method can be considered as a potential tool for obtaining a deeper understanding of the structure of data, with respect to a target property, in particular.
\end{abstract}

\section{Introduction}\label{introduction}
Situations where observation data are generated by different mechanisms in different contexts appear frequently in various scientific phenomena. Because of this reason, supervised machine learning models for predicting physical properties of materials still get frequently outmatched by empirical models created by human researchers. Finding a model that qualitatively determines the mixture effect is a constantly demanded task in both theoretical and experimental model construction. Use of unsupervised learning techniques, with the ability to screen predefined correlations at different data scales, can be a promising approach \cite{LeCun15}.  Conventional unsupervised learning techniques for unveiling of mixture models in descriptive space are implemented using clustering methods \cite{Xu2015} such as the Gaussian mixture model, hierarchical clustering, K-means clustering, etc. Besides, the unveiling of mixture models using supervised information, which considers functions connecting the descriptive space and target space as center objects, has not gained considerable attention from the machine-learning community. One of the well-known methods in this research direction is the mixture of experts model \cite{Jacobs91, Yuksel12}, which learns the gating functions to appropriately partition the descriptive space for identifying the components of mixture models. Further, linear regression-based clustering was recently developed \cite{Eto2014FullyAutomaticBP, NIPS2013_5171, Nguyen18}  without partitioning the descriptive space. However, these models are sensitive to parameters setting, including the number of clusters, complexity of the learners (linear model), etc. 

In this study, we propose a data-driven method to unveil the mixture of information on the mechanism of physical properties of materials by using nonlinear supervised learning techniques. We pay attention to extracting dissimilarity between materials, with respect to a given target physical property. The method is based on an ensemble method with Kernel ridge regression as the predicting model. We apply a bagging algorithm to carry out random subset samplings of the materials for generating multiple prediction models. The distribution of the predicted values for each material is then approximated by a Gaussian mixture model. Further, the contributions of the reference training materials to each of the corresponding models are investigated in detail. Reference training materials that are avoided and do not contribute to a predictive model, which accurately predicts the physical properties of a particular material, are considered dissimilar to that material. 

This paper is organized as follows: The components of the algorithm are described in Section \ref{Methodology}, including the base learning model in Section \ref{KRR}, the bagging algorithm in Section \ref{EnsembleLearning}, and the dissimilarity voting machine in Section \ref{VotingMachine}. Data preprocessing is elucidated in Section \ref{DataPreprocess}, and the results and discussions are presented in Section \ref{ResultDiscuss}.

\section{Methodology}\label{Methodology}
We consider a dataset $\mathcal{D}$ of $p$ materials. Assume that a material with index $i$ is described by an $m$-dimensional predictor variable vector, ${\bm{x}}_{i} = {\left( {x_i^1}, {x_i^2}, \dots , {x_i^m} \right)} \in {\mathbb{R}}^m$. The dataset $\mathcal{D} = \left\{ ({\bm x}_1, y_1), ({\bm x}_2, y_2) \dots ({\bm x}_p, y_p)\right\} $ is then represented using a $\left( p \times \left( m+1 \right) \right)$ matrix. The target physical property values of all materials in the dataset are stored as a $p$-dimensional target vector $ \bm{y} = \left( y_1, y_2 \dots y_p \right) \in \mathbb{R}^p$. 

\subsection{Kernel ridge regression}\label{KRR}
To learn a regression function $\hat{f}$ for predicting the target variable, we utilize the kernel ridge regression (KRR) technique \cite{ML}, which has been recently applied successfully in several materials science studies \cite{Rupp_tutorial, Botu, Pilania}. We use the KRR with Laplacian kernel function as follows:

\begin{equation}
k(\bm{x}_{i}, \bm{x}_{j})  =   \exp (\frac{-|\bm{x}_{i}-\bm{x}_{j}|}{\sigma})
\end{equation}
where $|\bm{x}_{i} - \bm{x}_{j}| = \sum_{a=1}^{m}|x_{i}^{a}-x_{j}^{a}|$ and $\sigma$ is the tuning variance parameter of the Laplacian kernel. 

For a given new material $\bm{x}_*$, the predicted property $\hat{f}(\bm{x}_*)$ is expressed as the weighted sum of the kernel functions:

\begin{equation}
\hat{f}(\bm{x}_*) = \sum_{i=1}^N c_i k(\bm{x}_*, \bm{x}_{i})
\label{eq.fx}
\end{equation}
where $N$ is the number of training materials. The weighting coefficients $c_i$ for the corresponding materials $\bm{x}_{i}$ are determined by minimizing

\begin{equation}
\sum_{i=1}^N [\hat{f}(\bm{x}_i) - {\bm y}_i]^2 + \lambda\sum_{i=1}^{N} ||c_i||_2^2.  
\label{eq.EL}
\end{equation}

The regularization parameter $\lambda$ and the hyper parameter $\sigma$ are selected by cross-validation\cite{Stone74, Picard84}, i.e., by excluding some of the materials during the training process, and maximizing the prediction accuracy for those excluded materials.  We consider the component $c_i k(\bm{x}_*, \bm{x}_{i})$ in Eq. \ref{eq.fx} as the contribution of the training material $\bm{x}_{i}$ to the prediction model $\hat{f}$ and if it is not refered in the training set,  $c_i$ is set to zero.

\subsection{Testing for homogeneity of dataset with ensemble learning}\label{EnsembleLearning}
Ensemble learning \cite{Hastie09, Dietterich2000} is a method in machine learning where multiple learners are trained to solve the same problem, which was initially developed to avoid over fitting for our designed base learner \cite{zhou2012}. There are various strategies, such as bagging, boosting, and stacking, for different learning purposes.

In this study, we intend to unveil the mixture of information in prediction model space, which is a linear combination of kernel functions constructed by training materials. Applying the bagging algorithm\cite{BALDI201478}, we carry out random subset samplings of the materials dataset to generate multiple prediction models. For each sampling, we prepare two datasets:  bagging dataset, $\mathcal{D}_{bagg}$, and testing dataset, $\mathcal{D}_{test}$. These two datasets satisfy $\mathcal{D}_{bagg} \cap \mathcal{D}_{test} = \emptyset$ and $\mathcal{D}_{bagg} \cup \mathcal{D}_{test} = \mathcal{D}$. With each of the two datasets ${\mathcal{D}_{bagg}, \mathcal{D}_{test}}$, we generate a prediction model by regressing the bagging datasets $\mathcal{D}_{bagg}$ using a cross-validation technique \cite{Stone74, Picard84}. For each obtained prediction model, we collect the predicted values of the target property for all the materials in the corresponding testing dataset $\mathcal{D}_{test}$. The canonical size of $\mathcal{D}_{bagg}$ is selected as 66\% of the total number of data instances. By repeating the bagging process, each material $\bm{x}_i$ has an equal chance to appear in the test set $\mathcal{D}_{test}$. Finally, we can obtain a distribution of the predicted values of target property $p(\hat{y}(\bm{x}_i))$ for all materials.

Here, the null hypothesis stands for an assumption of the homogeneity of the dataset in the kernel space or the existence of a single regression function. If the null hypothesis is true, the distribution $p(\hat{y}(\bm{x}_i))$ should be Gaussian for every material $\bm{x}_i$; else, we can significantly approximate the distribution $p(\hat{y}(\bm{x}_i))$ for a particular material $\bm{x}_i$ in the form of a mixture of Gaussian distributions. By examining the distribution $p(\hat{y}(\bm{x}_i))$,  we can test the hypothesis on the homogeneity of our dataset. 

The approximation of the distribution $p(\hat{y}(\bm{x}_i))$ by a mixture model of $K$ Gaussian distributions \cite{GMM} is as follows:
\begin{equation}
p(\hat{y}(\bm{x}_i)|\theta) = \sum_{k=1}^{K} {\pi}_{i}^{k} \mathcal{N}({\mu}_{i}^{k}, {\sigma}_{i}^{k}),
\label{eq.GMM}
\end{equation}
where ${\pi}_{i}^{k}$, ${\mu}_{i}^{k}$, and ${\sigma}_{i}^{k}$ are the weights, centers, and coefficient matrices of the constituent Gaussians components. For a given number of mixture components, the parameters are estimated using an expectation-maximization algorithm, which is explained in detail in \cite{GMM}. To determine the number of mixture components, a maximizing Bayesian information criterion \cite{SchwarzBIC} process is utilized by applying several different trials to randomize the initial states. 

\subsection{Dissimilarity voting machine}\label{VotingMachine}
In this section, we utilize the information from the bagging experiment to vote for the dissimilarity among materials.  To perform the dissimilarity voting procedure, first, under a predefined tolerance in prediction error $\delta_{thres}$, all prediction models $\hat{f}$ learned from a dataset $\mathcal{D}_{bagg}$ (Eq. \ref{eq.fx}), which satisfies $|\hat{f}(\bm{x}_{i})-y_i|< \delta_{thres}$, are collected. Then, under a predefined neighborhood condition in description space $k_{thres}$, for a given $\bm{x}_{j}$ in the $\mathcal{D}_{test}$ and all $\bm{x}_{i}$ in {$\mathcal{D} - \{x_j\}$, a vote $ds(\bm{x}_{i}, \bm{x}_{j})$ for the dissimilarity between $\bm{x}_{i}$ and $\bm{x}_{j}$ is defined:

\begin{equation}
ds(\bm{x}_{i}, \bm{x}_{j}) =
	\begin{cases}
      1, & \text{if}\ c_{i} = 0 \ and \  k(\bm{x}_{i}, \bm{x}_{j}) < k_{thres}\\
      0, & otherwise
    \end{cases}.
\label{eqn.voting}
\end{equation}

In this voting machine, for each material, we pay more attention on its relationship with the neighborhood materials (in the data set) in the description space. If the neighborhood materials are avoided and make no contribution to the predictive models that accurately predict the physical property value of the concerning material, those neighborhood materials should be considered dissimilar to that material. Finally, the bagging-based dissimilarity voting algorithm is summarized as follows: 
\begin{algorithm}
\caption{Bagging-based dissimilarity voting algorithm}\label{algorithm}
\KwData{ 
	\begin{description}
		\item \makebox[4.2cm][l]{Dataset:} $\mathcal{D} = \left\{ ({\bm x}_1, y_1), ({\bm x}_2, y_2) \dots ({\bm x}_p, y_p)\right\} $ 
		\item \makebox[4.2cm][l]{Base learning:} $\hat{f}$
		\item \makebox[4.2cm][l]{Number of base learners:} $T$
	\end{description}
}
\Parameter{ $k_{thres}, \delta_{thres}$}
 \KwResult{Dissimilarity matrix, $\bm{dS}$}
 
 \SetKwBlock{Begin}{begin}{End}
\Begin{
	\For{$t\gets1$ \KwTo $T$ \KwBy $1$}{
		$h_t = \hat{f}(\mathcal{D}, \mathcal{D}_{bagg})$ \\
	}
	$H({\bm x}) = \sum_{t=1}^{T} \mathbb{I}\left(h_{t}({\bm x}_{*}) \leq \delta_{thres}\right)$ \\
	${\bm{dS}} = 0$ \text{with} ${\bm{dS}} = {\left[ ds_{ij} \right]}_{p \times p}$ \\
	\ForEach{$h_t \in H$}{
		\ForAll{ $k({\bm x}_{i}, {\bm x}_{*t}) \leq k_{thres}$ }{
		
				\uIf{ $c_{i} = 0$}{
					${ds_{ij}} \mathrel{+}= 1\ \forall {\bm x}_{j} \in h_t $
				}
		}
	}
}
\Return ${\bm{dS}}$.

\end{algorithm}

\section{Data collection and representation}\label{DataPreprocess}
\subsection{Prototype model} \label{DataToymodel}
We simulate a dataset containing 70 data instances with a one-dimensional descriptive variable, $x$, and target variable, $y$, as shown in Figure \ref{fig.ToyModel}a. To illustrate the capability of the bagging prediction model in unveiling the mixture of nonlinear models, the dataset is designed as a mixture of three main functions. In the range of $x$ lesser than -0.4, the function $y = f(x)$ is monotonic and centered at 0.1. In the range of $x$ greater than -0.4, the function $f$ is bifurcated with a branch fluctuation increasing from 0.1 to 0.3 and the other variation decreasing from 0.1 to -0.15. 
\begin{figure}[t]
\centering
\includegraphics[width=1.0\textwidth]{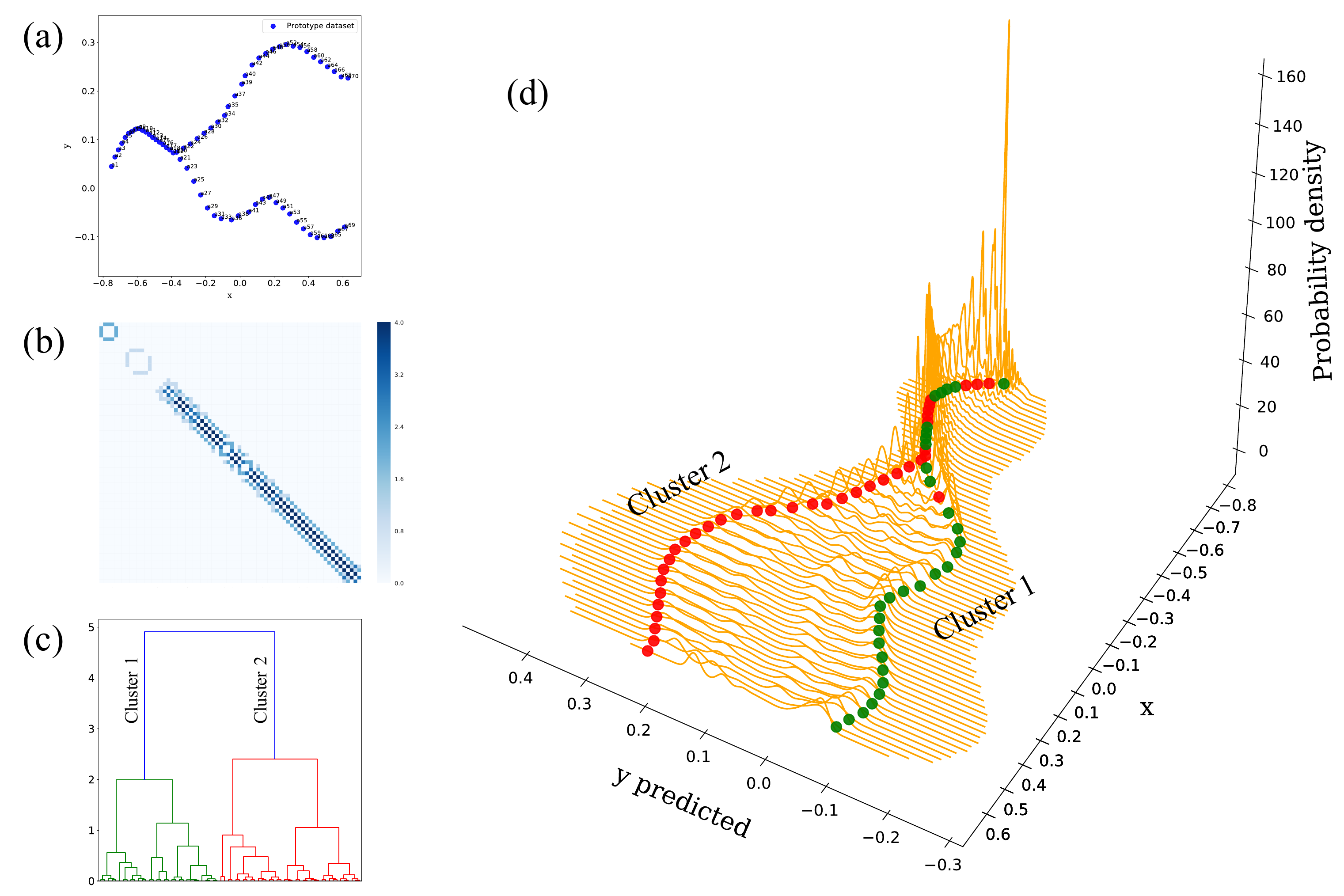}
\caption{  a) Visualization of the prototype data with the one-dimensional predictor variable, $x$, and target variable, $y$. b) Dissimilarity voting matrix with colored cells show the dissimilarity pairs of materials. Dark blue cells indicate pairs of data instances associated with high dissimilarity voting information and vice versa. c) Hierarchical clustering graph is constructed by embedding information of dissimilarity voting matrix. d) Distribution of the predicted values $y$ along the $x$ axis, applying the bagging model (orange lines) and observed data (green and red points), which are clustered using hierarchical clustering technique and information from dissimilarity voting results.} \label{fig.ToyModel}
\end{figure}

\subsection{Curie temperature data} \label{DataTc}

We collected experimental data on 101 binary materials, consisting of transition metals and rare-earth metals, from the Atomwork database of NIMS \cite{paulingfile, atomwork}, including the crystal structure of the materials and their observed {\TC} values. Our task was to develop a model for estimating the {\TC} of a new material based on the training data of known materials. For this, one of the crucial steps is the selection of an appropriate data representation that reflects the application domain, i.e., a model of the underlying physics. On the other hand, data representation that derives a good estimation model may imply the discovery of the underlying physics. To represent the structural and physical properties of each binary material, we designed 21 descriptive variables. We divided the 21 variables into three groups; the first and second categories contained descriptive variables that describe the atomic properties of the transition-metal ($T$) and rare-earth ($R$) constituents. The properties were as follows: (1, 2) atomic number ($Z_R$, $Z_T$), (3, 4) covalent radius ($r_{covR}$, $r_{covT}$), (5, 6) first ionization (${IP}_R$, ${IP}_T$), and (7, 8) electronegativity (${\chi}_R$, ${\chi}_T$). In addition, descriptive variables related to the magnetic properties were included, as follows: (9, 10) total spin quantum number ($S_{3d}$, $S_{4f}$), (11, 12) total orbital angular momentum quantum number ($L_{3d}$, $L_{4f}$), and (13, 14) total angular momentum ($J_{3d}$, $J_{4f}$). 

The selection of these variables originates from the physical consideration that the intrinsic electronic and magnetic properties determine the 3$d$ orbital splitting at the transition-metal sites. To better capture the effect of the $4f$ electrons, the strong spin-orbit coupling effect in particular, we included three additional variables for describing the properties of the constituent rare-earth metal ions: the (15) projection of the total magnetic moment onto the total angular moment ($J_{4f}g_{j}$), and (16) projection of the spin magnetic moment onto the total angular moment ($J_{4f}\left(1-g_{j}\right)$) of the 4$f$ electrons. The selection of these variables was based on the physical consideration that the magnitude of the magnetic moment determines {\TC}. It has been well established that information related to the crystal structure is highly important for understanding the physics of binary materials with transition metals and rare-earth metals. Therefore, we designed a third group with structural variables that approximately represent the structural information at the transition metal and rare-earth metal sites, including the (17) concentration of the transition metal ($C_T$), and (18) concentration of the rare-earth metal ($C_R$). Note that, if we use the atomic percentage for the concentration, the two quantities are not independent. Therefore, in this study, we measured the concentrations in units of atoms/\AA$^3$. This unit is more informative than the atomic percentage because it contains information on the constituent atomic size. As a consequence, $(C_T)$ and $(C_R)$ are not completely dependent on each other. Other structure variables were also included: the mean radius of the unit cell (19) between two rare-earth elements, $r_{RR}$, (20) between two transition metal elements, $r_{TT}$, and (21) between transition and rare-earth elements, $r_{TR}$. We set the experimentally observed {\TC} as the target variable.

\section{Results and discussion}\label{ResultDiscuss}
\subsection{Prototype model} \label{ResultToymodel}

To demonstrate the effect of applying the bagging prediction model in investigating the structural insight dataset, we present the results of applying the model to two-dimensional simulation data. The dataset contains 70 instances with a one-dimensional descriptive variable, $x$, and target variable, $y$, as depicted in Figure \ref{fig.ToyModel}b and Section \ref{DataToymodel}. The bagging prediction model includes one million random samplings, with the sampling size is 35\% of the total number of data instances. Details about setting parameters are described in Table 1 in Supplemental Materials.

Figure \ref{fig.ToyModel}a shows the distributions of the predicted values, ${\hat y}$, obtained using the bagging model. It is obvious that, for $x$ values lesser than -0.4, the ${\hat y}$ distributions include a single distribution centered around 0.1. On the other hand, for $x$ values greater than -0.4, almost all the ${\hat y}$ distributions can be considered to be a mixture of two main Gaussians, whose mean are approximately 0.2 and -0.1, respectively. These distribution components of the predicted value, ${\hat y}$ reflect the actual shape of the designed data, shown by colored points. 

Figure \ref{fig.ToyModel}b shows the dissimilarity $(70 \times 70)$ matrix obtained by our developed dissimilarity voting machine. In the matrix, dark blue cells represent dissimilarity pairs of data instances. Zero value cells show no dissimilarity information between corresponding pairs. For convenience, the ordered data instances shown in the matrix are sorted by the $x$ values. Details about how the voting machine works to detect dissimilarity effect are shown by zero contribution example profiles in Supplemental Materials.

There are two noticeable points extracted from this figure. First, the upper left of the matrix represents a large bright region or the region of non-dissimilarity among instances. It is consistent with the monotone and smoothly changes for $x$ values lesser than -0.4. Second, for $x$ values larger than -0.4, one can notice that any data instances in this region are dissimilar with their two closest neighbors and similar to the next ones. Once again, this extracted information shows consistency with the actual distribution of the designed dataset. 

By transferring information from the dissimilarity matrix to hierarchical clustering model \cite{hac11}, we obtain clustering results as shown in figure \ref{fig.ToyModel}c. The dissimilarity information helps us to identify a mixture of two main groups, which are labeled by green and red. These two groups successfully reconstruct the shape of two branches in the bifurcated region of the dataset. To conclude, through the prototype model result, the dissimilarity voting machine based on the bagging algorithm is shown to possess the ability to identify the mixture phenomena regarding a specific target property.

\begin{figure*}[t]
\centering
\subfloat[][]{\includegraphics[width=0.33\textwidth]{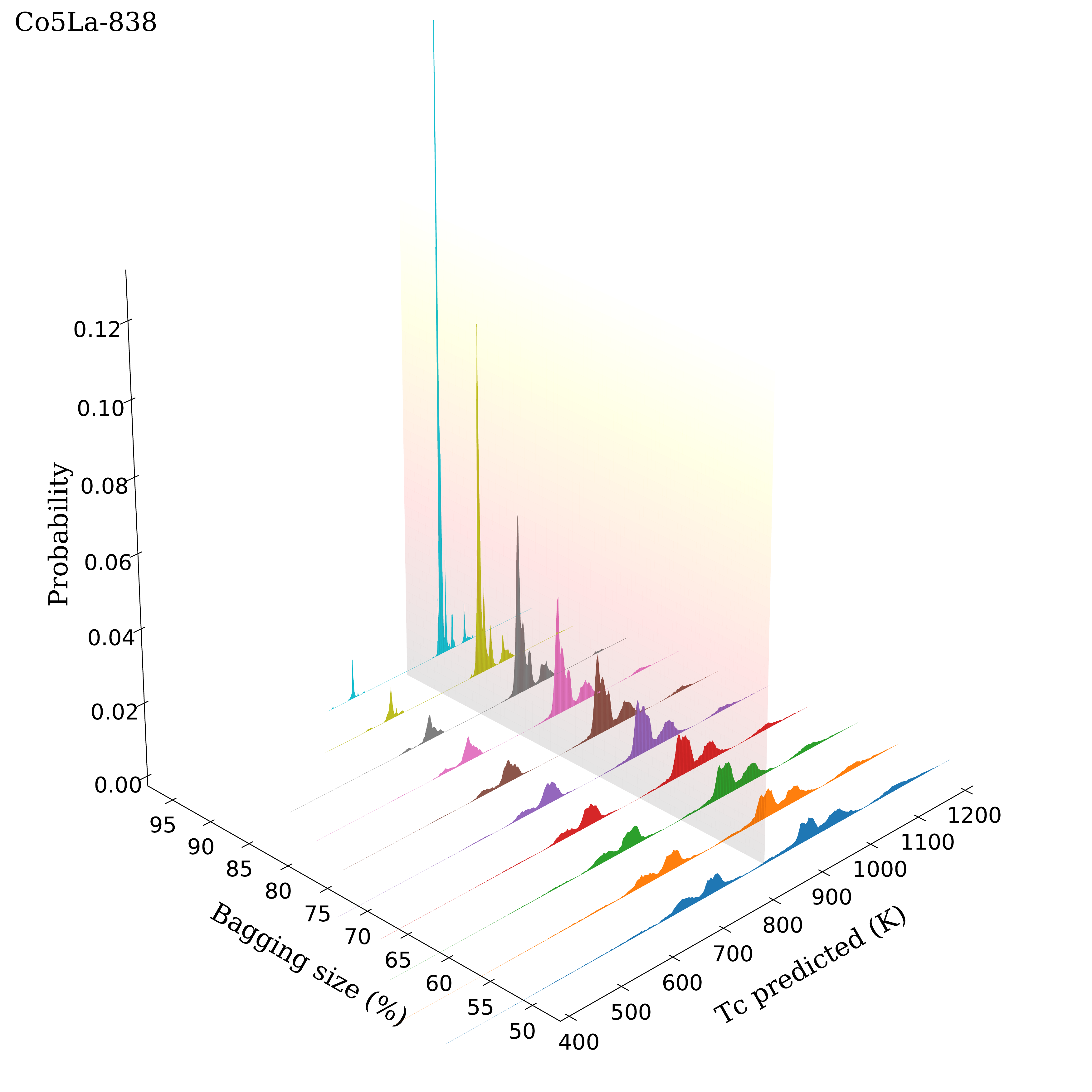}}
\subfloat[][]{\includegraphics[width=0.33\textwidth]{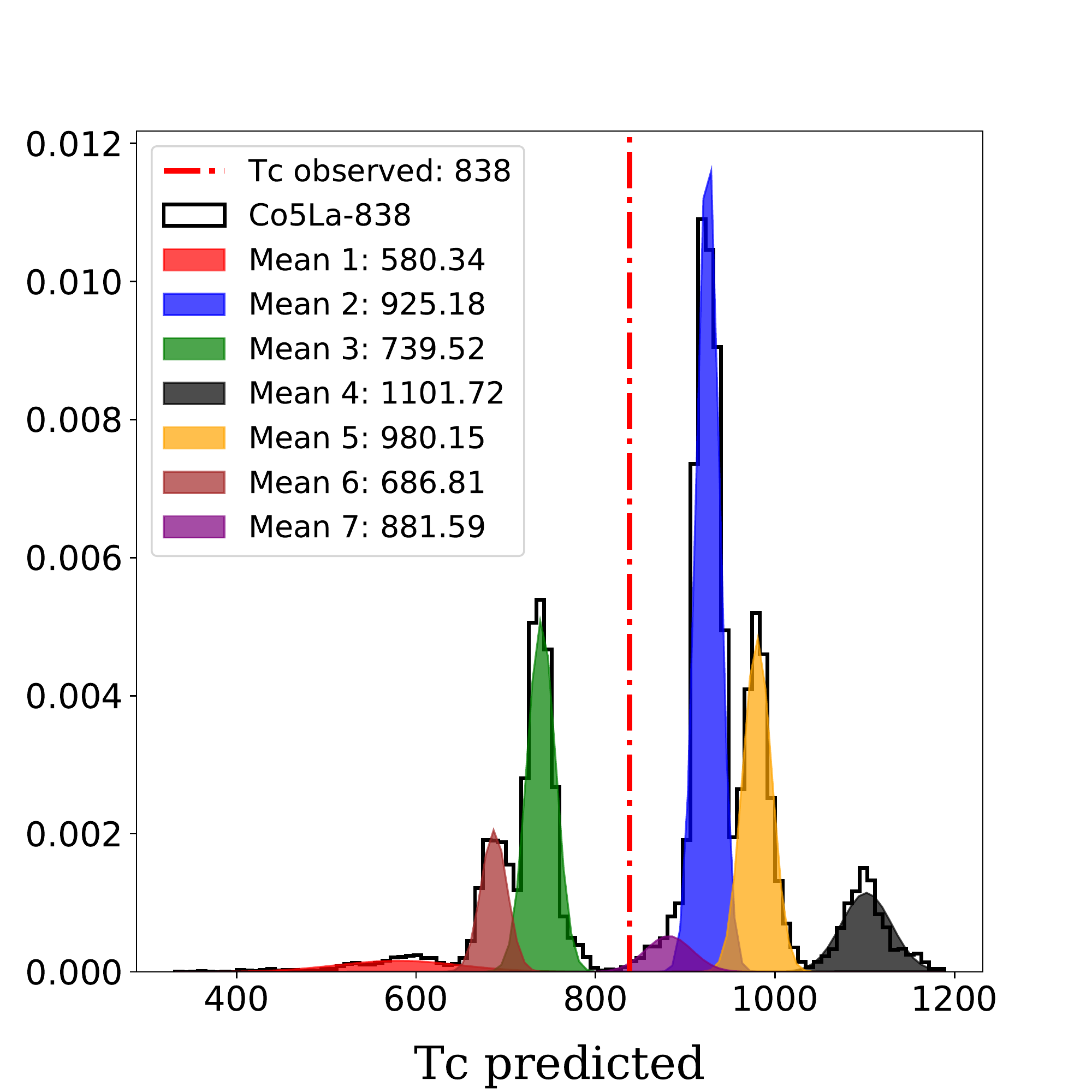}}
\subfloat[][]{\includegraphics[width=0.33\textwidth]{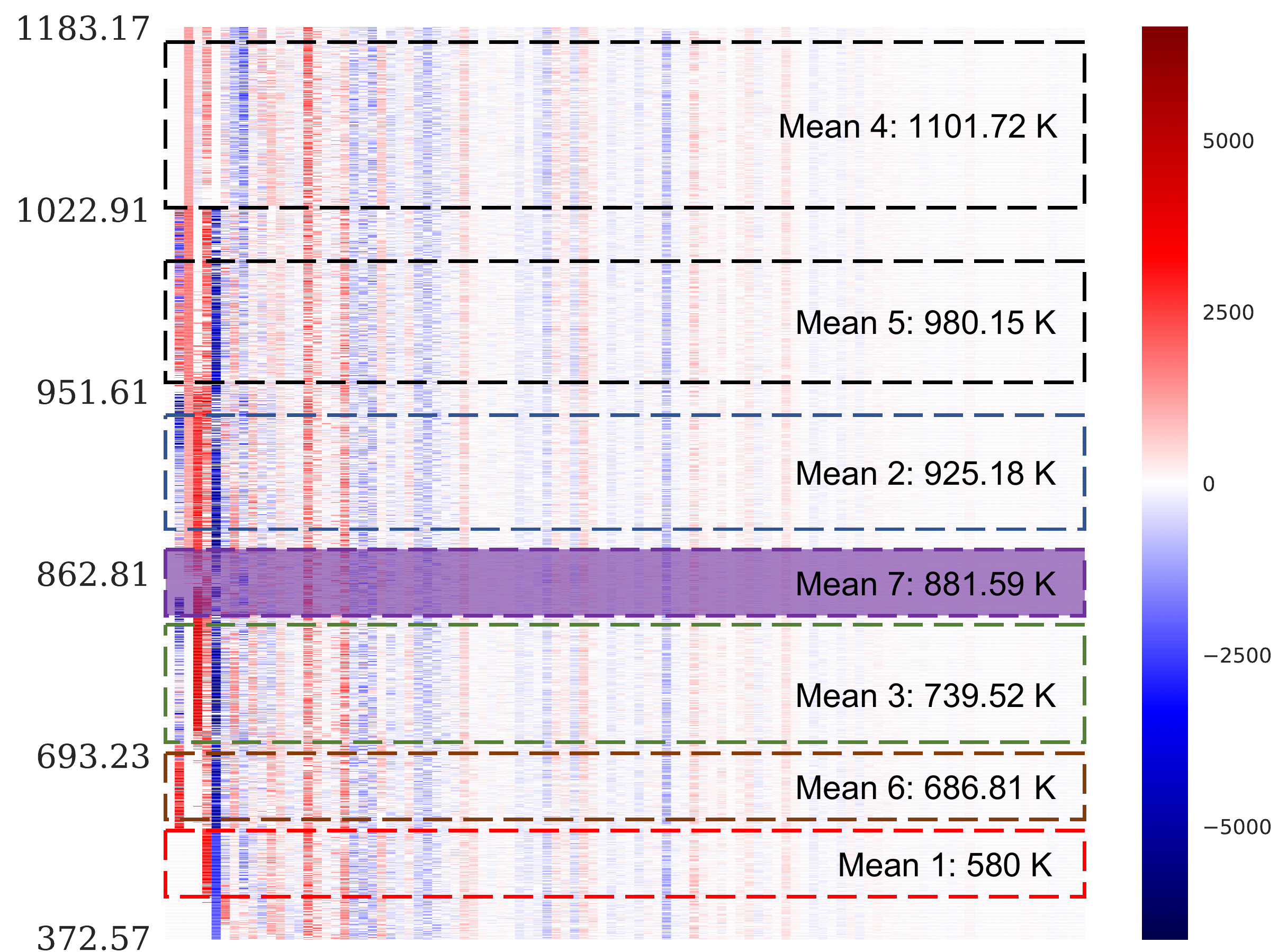}}
\caption{a) {\TC} predicted-value distribution of $\rm Co_{5}La$ for different bagging sizes. The constant plane shows the position of the observed value. b) {\TC} predicted-value distribution of $\rm Co_{5}La$ with a bagging size of 65\% of the total data instances in the dataset. The distribution is a mixture of seven Gaussian components. The red dashed lines indicate the positions of the observed values. c) Heat map matrix depicting the contribution of the training materials to the target material for the different learners in Figure b). The horizontal axis represents training materials sorted by the L1 distance to the target material on the description space. The vertical axis represents those sorted by the predicted {\TC} value, i.e., the summation of all the contributions.}
\label{fig.PredictedDistTest}
\end{figure*}

\subsection{Curie temperature analysis} \label{ResultTc}

For the designed descriptive variables, the maximum prediction ability with an $R^2$ score of $0.967 \pm 0.004$ was achieved by a model derived from the variable combinations, $\left\{ {\chi}_R, {\chi}_T, J_{4f}\left(1-g_{j}\right), Z_T, r_{covT}, {IP}_T, S_{3d}, L_{3d}, J_{3d}, C_R\right\}$. Details about setting parameters are described in Table 1 in Supplemental Materials. The model selection and relation among variables are discussed in \cite{Dam2018, Nguyen2018}. The high prediction accuracy level of this model shows that it is possible to accurately predict the {\TC} values of rare-earth transition bimetal materials with the designed variables. With the bagging size at level of 90\% total data instances in the data set, the mean absolute value of cross-validation is approximate 40K. The summation about the bagging size dependence of mean absolute error in cross-validation process is shown in Figure 2 in Supplemental Materials. Details about training-testing prediction errors for all materials is shown in Figure 1 in Supplemental Materials. From these evidences, under the description of this variable combination, the regression function to predict {\TC} is approximately a single function associated with a number of probable unknown anomalies. Our designed dissimilarity voting machine could help to address this problem.

Next, we discuss the distribution of the predicted {\TC}. Almost all materials obtained its own predicted values follow a single Gaussian function distribution, whereas some materials were associated with distributions that were a mixture of Gaussians. For example, figure \ref{fig.PredictedDistTest}a shows the {\TC} predicted-value distribution of $\rm Co_{5}La$ (with observed {\TC} of 838~K) for bagging sizes varying from 50--95 percent of the total dataset. It is clear that for all values of the size in the subset sampling selection, there is a consistent form of distribution. On changing the size, the corresponding peaks at 686~K, 739~K, 925~K, and 980K remained. Figure \ref{fig.PredictedDistTest}b displays an enlarged view of the distribution for a bagging size of 65 percent of the total dataset. The detailed result, on applying the Gaussian mixture model (Sect. \ref{EnsembleLearning}), shows that this distribution is a mixture of seven Gaussian distributions, whose means are at 580.34~K, 686.81~K, 739.52~K, 881.59~K, 925.18~K, 980.15~K, and 1101.72~K, respectively. This indicates the presence of a mixture of nonlinear functions in the structural insight dataset. Further investigation will reveal the significance of appearance of these functions.

Figure \ref{fig.PredictedDistTest}c shows the contribution of each training material in the dataset to the target material, $\rm Co_{5}La$, in detail. The color bar shows the color encoding of the contributions, in Eqn. \ref{eq.EL} for all materials in the dataset to $\rm Co_{5}La$. It is zero-center symmetrical color encoding with the white color corresponds to non referring materials in training set. The vertical axis represents those sorted by the predicted {\TC} value, i.e., the summation of all the contribution values as in Eqn. \ref{eq.EL}. The horizontal axis represents materials with an ascending order of the $L_{1}$ distance to the target material. The top five closest to $\rm Co_{5}La$ are: $\rm Co_{5}Ce~(662~K)$, $\rm Co_{13}La~(1298~K)$, $\rm Co_{17}Ce_{2}~(1090~K)$, $\rm Co_{5}Pr~(931~K)$,and $\rm Co_{7}Ce_{2}~(50~K)$. This figure also shows that models with the closest predicted value of {\TC} (purple distribution with a mean of 881.59~K) are constructed from a combination of instances, $\mathcal{D}_{bagg}$, with no contribution from $\rm Co_{5}Ce$, $\rm Co_{13}La$, $\rm Co_{17}Ce_{2}$, and $\rm Co_{7}Ce_{2}$. Only $\rm Co_{5}Pr$ shows significant contribution to $\rm Co_{5}La$. Details about zero-contribution counting profiles are shown in Figure 4 in Supplemental Materials section.

For comparison, the two nearest neighbor models of the purple model, namely, the green model with mean 739~K, and blue model with mean 925.18~K were considered. For the blue model, all the five nearest neighbors were considered as contributors, and in the green model, the contribution of $\rm Co_{13}La$ was removed. The results from the training data contributions, shown above, provide certain conclusions on the actual physical meaning. The three materials, $\rm Co_{5}Ce$, $\rm Co_{13}La$, and $\rm Co_{17}Ce_{2}$, should not be considered highly "similar" to $\rm Co_{5}La$ with respect to $T_C$, even though they have the same constituent $T$-metal and the same constituent $R$-metals or are positioned next to each other on the periodic table ($Z_{La}$=57 and $Z_{Ce}$=58). In other words, the distance between these materials with respect to the rare-earth element difference should not be close, as measured by the three $R$ predictive variables, $\chi{R}$, $J_{4f}(1 - g_{j})$, and $C_{R}$. As the concentration of the $R$-metal efficiently indicates the change in {\TC} values among those sharing the same $R$ and $T$ materials, e.g., $\rm Co_{5}La$ vs $\rm Co_{13}La$, these dissimilarity results show that the other two $R$ variables do not capture the real mechanism of {\TC}. 
\begin{figure}[t]
\centering
\subfloat[][]{\includegraphics[width=0.5\textwidth]{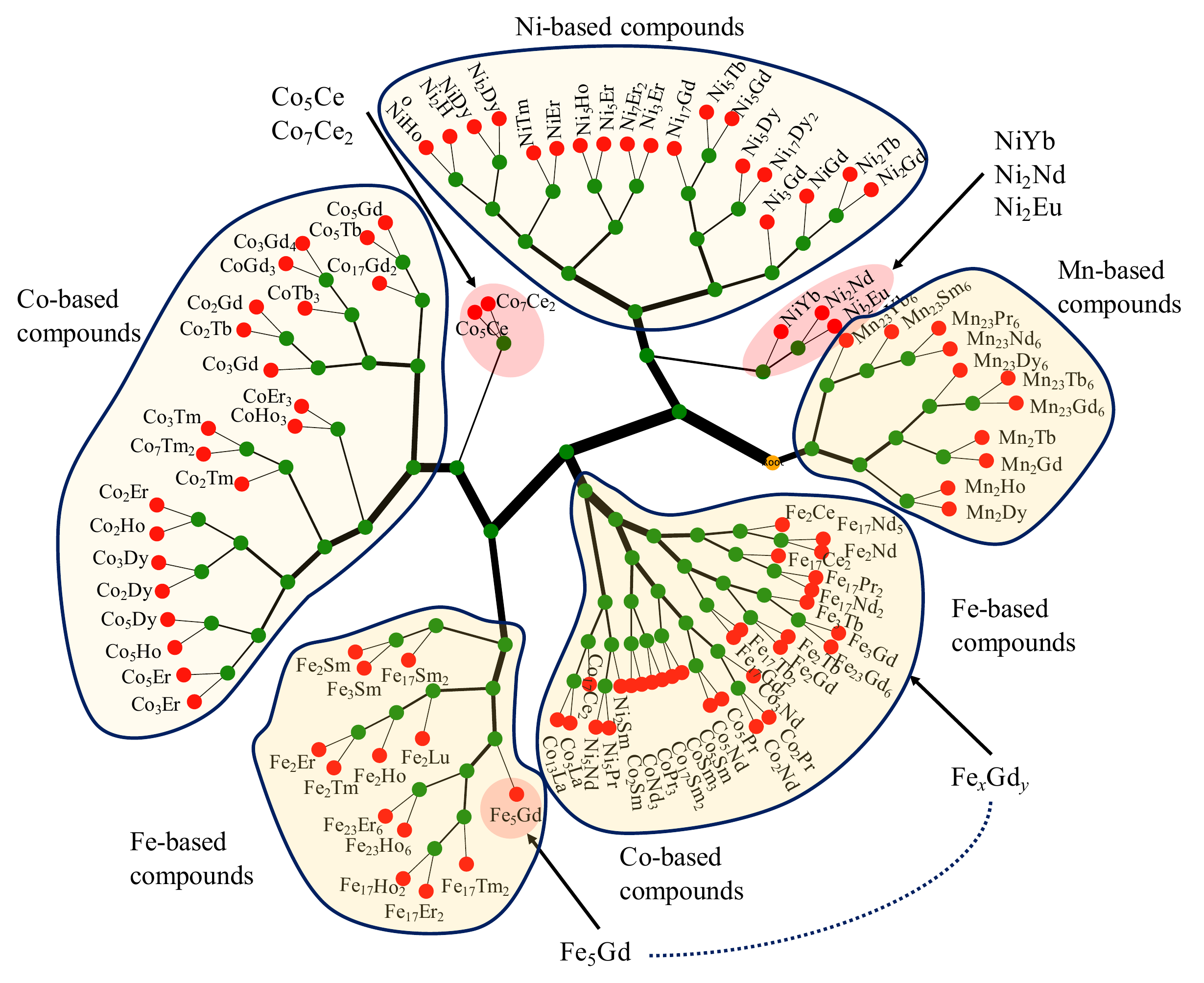}\label{fig.TcHac}}
\subfloat[][]{\includegraphics[width=0.5\textwidth]{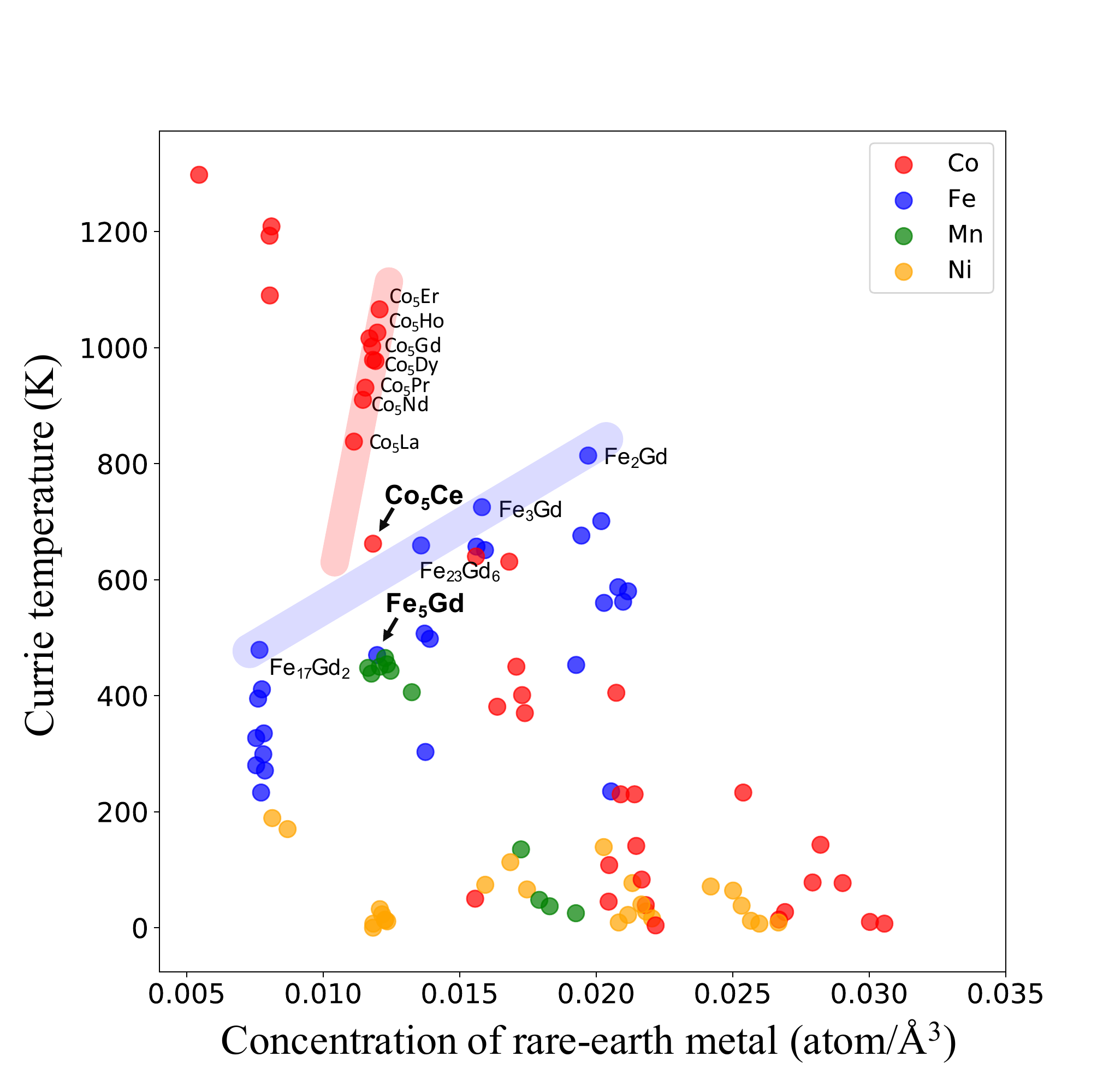}\label{fig.Tc_CR}}
\caption{a) Hierarchical clustering model by utilizing information from dissimilarity matrix. b) Relationship between {\TC} and the concentration of rare-earth element, $C_R$. Red and blue lines show anomalies detection which is discussed in detail in Section \ref{ResultTc} }
\end{figure}

The entire dataset can be divided into four main groups based on transition metals: cobalt-based, iron-based, manganese-based, and nickel-based materials. In our defined $k_{thres}$ neighbor regions, the number of dissimilarity values was not identical for all the groups of materials. Here, we analyze the cobalt-based and iron-based material groups. In the cobalt-based group, we can notice that $\rm Co_{5}Ce$ does not receive contributions from the other materials of the group $\rm Co_{5}R$. The dissimilarity between the $\rm Co_{5}Ce$ and the $\rm Co_{5}La$ material was shown in the previous analysis. Here, the dissimilarity can be observed more distinctly. Compared to the other $\rm Co_{5}R$ materials, $\rm Co_{5}Ce$ has a {\TC} of 662~K, which is considerably lower than those of $\rm Co_{5}La$ at 838~K, $\rm Co_{5}Pr$ at 931~K, $\rm Co_{5}Nd$ at 910~K, and $\rm Co_{5}Sm$ at 1016~K. In this family, except for $\rm Co_{5}Ce$, an increase in the atomic number of the rare-earth element correlates to an increasing {\TC} value. Figure \ref{fig.TcHac} shows the hierarchical clustering result obtained by utilizing the information from the dissimilarity voting machine. It is obvious that $\rm Co_{5}Ce$ is isolated from other Co-based materials. We can also confirm the anomalousness of $\rm Co_{5}Ce$ by comparing $\rm Co_{5}Ce$ with other Co-based materials (Fig. \ref{fig.Tc_CR}, the compounds surrounded by blue line). This result confirms the significance of our method of dissimilarity measurement.

In the group of iron-based materials, Figure. \ref{fig.TcHac} shows that $\rm Fe_{5}Gd$ appears with a large number of dissimilarity values compared to other Fe-based materials -- especially the $\rm Fe_{x}Gd_{y}$ group, indicating that $\rm Fe_{5}Gd$ is out of trend with its nearest neighbors.  From Figure \ref{fig.Tc_CR} (the compounds surrounded by blue line), it can be shown that, for an increasing concentration of rare-earth elements {$C_R$}, $\rm Fe_{17}Gd_{2}$, $\rm Fe_{5}Gd$, $\rm Fe_{23}Gd_{6}$, $\rm Fe_{3}Gd$, and $\rm Fe_{2}Gd$, the {\TC} values are 479~K, 465~K, 659~K, 725~K, and 814~K, respectively. It is clear that, $\rm Fe_{5}Gd$ does not follow the general trend of $\rm Fe_{x}Gd_{y}$ groups. Thus, it is again demonstrated that the information extracted by dissimilarity voting using the bagging algorithm could be used as a useful method to detect anomalies.

\section{Conclusion}
In this study, we have proposed a method to extract dissimilarity between materials, with respect to a given target property. This technique is based on an ensemble method with Kernel ridge regression as the predicting model; multiple random subset sampling of the materials is performed to generate prediction models and corresponding contributions of the reference training materials in detail. The mixture distribution of predicted values was unveiled using a Gaussian mixture models. The reference training materials contributed to the prediction model that accurately predicts the physical property value of a specific material, are considered to be similar to that material, or vice versa. Evaluations using synthesized data demonstrate that the proposed method can effectively measure the dissimilarity between data instances. Next, the algorithm was applied for analyzing the Curie temperature ({\TC}) prediction of the binary 3$d$ transition metal - 4$f$ rare-earth binary alloy problem and exhibited satisfactory results. The proposed method can be considered as a potential tool for obtaining a deeper understanding of the structure of data, with respect to a target property, in particular.

\section*{Acknowledgments}
This work was partly supported by PRESTO and the ``Materials Research by Information Integration'' initiative (MI$^2$I) project of the Support Program for Starting Up Innovation Hub, by the Japan Science and Technology Agency (JST) and the Elements Strategy Initiative Project under the auspices of MEXT, and also by MEXT as a social and scientific priority issue (Creation of New Functional Devices and High-Performance Materials to Support Next-Generation Industries; CDMSI) to be tackled using a post-K computer.

\section*{References}
\bibliography{main}

\end{document}